\documentclass[onecolumn]{clusterlab_paper}
\usepackage{graphicx}
\usepackage{tablefootnote}
\usepackage{footnote}
\usepackage{import}
\usepackage{multirow}
\usepackage{fancyhdr}

\usepackage{hyperref}
\usepackage{color}
\usepackage{makecell}
\usepackage{adjustbox}
\usepackage{amssymb}
\usepackage{makecell, cellspace, caption}
\usepackage{array}
\newcolumntype{L}[1]{>{\raggedright\let\newline\\\arraybackslash\hspace{0pt}}m{#1}}
\newcolumntype{C}[1]{>{\centering\let\newline\\\arraybackslash\hspace{0pt}}m{#1}}
\newcolumntype{R}[1]{>{\raggedleft\let\newline\\\arraybackslash\hspace{0pt}}m{#1}}
\usepackage[toc,page,title,titletoc,header]{appendix} 
\usepackage[english]{babel}
\usepackage{array}
\newcolumntype{P}[1]{>{\centering\arraybackslash}p{#1}}
\newcolumntype{M}[1]{>{\centering\arraybackslash}m{#1}}
\usepackage{amsmath}
\usepackage{amssymb}
\usepackage{stackengine}
\usepackage{comment}
\usepackage{pdfpages}
\usepackage{wrapfig}
\setcitestyle{number,square}
\usepackage{xspace}  
\usepackage{lipsum}
\usepackage{blindtext}
\usepackage{array}
\usepackage{tabularx}
\usepackage{longtable}
\usepackage{xltabular}
\usepackage{placeins}
\usepackage{breqn} 
\usepackage{makecell}
\usepackage{multirow} 

\usepackage[framemethod=TikZ]{mdframed}
\usepackage{tcolorbox}
\usepackage{multicol}
\usepackage{textcomp}
\usepackage{arydshln}

\newcommand{\authorfootnote}[1]{%
  \begingroup
  \renewcommand\thefootnote{\fnsymbol{footnote}}%
  \footnotetext[1]{#1}%
  \endgroup
}

\title{GemmAr: Enhancing LLMs Through Arabic Instruction-Tuning}

\multiauthors

\author{
    name={Hasna Chouikhi},  affiliation={1}
}
\author{
    name={Manel Aloui}, affiliation={1}
}

\author{
    name={Cyrine Ben Hammou}, affiliation={2}\textsuperscript{*}}

\author{
    name={Ghaith Chaabane}, affiliation={1}
}
\author{
    name={Haithem Kchaou}, affiliation={1}
}
\author{
    name={Chehir Dhaouadi},affiliation={1}
    
}

\corresponding[*] {\{hasna,manel,ghaith,haithem,chehir\}@clusterlab.com \\ \{c.benhamoud18240\}@pi.tn}
\affiliations{$^1$Clusterlab Team\\ $^2$Polytech INTL}

\abstract{

Large language models (LLMs) have greatly impacted the natural language processing (NLP) field, particularly for the English language. These models have demonstrated capabilities in understanding and generating human-like text. The success of language models largely depends on the availability of high-quality instruction datasets, which consist of detailed task descriptions and corresponding responses that are essential for training the models to address a variety of prompts accurately. However, the availability and quality of these resources vary by language. While models perform well in English, they often need help with languages like Arabic, due to the lack of datasets for fine-tuning Arabic-specific tasks. To address this issue, we introduce \textbf{InstAr-500k}, a new Arabic instruction dataset created by generating and collecting content that covers several domains and instruction types. We assess this dataset by fine-tuning an open-source Gemma-7B model on several downstream tasks to improve its functionality.
Based on multiple evaluations, our fine-tuned model achieves excellent performance on several Arabic NLP benchmarks. These outcomes emphasize the effectiveness of our dataset in elevating the capabilities of language models for Arabic. Our instruction dataset bridges the performance gap between English and Arabic language models by providing resources that amplify Arabic NLP development. Building on this foundation, we developed a model, \textbf{GemmAr-7B-V1}, specifically tuned to excel at a wide range of Arabic NLP tasks.

}

\begin{document}
\authorfootnote{Work done during the internship at Clusterlab.}
\section{Introduction}
The emergence of large language models (LLMs) has significantly developed the field of language technologies \cite{77,78}. These models exhibit capabilities in natural language understanding and generation. A fundamental aspect of improving these models involves instruction-tuning, where LLMs are trained on input/output pairs to refine their ability to follow specific user instructions.
\newline
Instruction-tuning has been extensively developed for English, contributing significantly to advancements in language technologies through improved model performance and understanding. Previous studies have highlighted the effectiveness of this approach in improving models' knowledge and reasoning capabilities \cite{intro2}. Despite these achievements, there is a significant disparity in the focus on languages other than English, particularly Arabic \cite{intro1}. This gap is important, especially given the digital expansion and the increasing demand for Arabic language technologies.
\newline
We aim to bridge this gap by developing instruction datasets for Arabic and adapting non-Arabic open-source models, such as Gemma-7B-IT \cite{gemma}, for Arabic-specific applications. This initiative aims to improve the accessibility of AI technology for Arabic speakers and contribute to more inclusive technological progress. We initiate our approach by creating an Arabic dataset (\textbf{InstAr-500k}) and then fine-tuning open-source LLM models to function effectively in Arabic. Our methodology involves synthetic data generation, human-crafted data collection, and using the LoRA technique \cite{lora} within the LLaMAFactory framework \cite{llamafact} for fine-tuning. The effectiveness of these models is assessed against Arabic NLP benchmarks using evaluation metrics to validate our methods. 
\newline
By pursuing these efforts, we successfully developed a new model \textbf{GemmAr-7B-V1} that demonstrates capabilities in handling a variety of tasks related to the Arabic language. This model was specifically fine-tuned to address the unique syntactic and semantic complexities of Arabic. \newline 
Its evaluation scores on various Arabic NLP benchmarks quantitatively reflect the effectiveness of \textbf{GemmAr-7B-V1}. These benchmarks encompass a range of tasks designed to test the models' understanding of context, nuance, and the intricacies of Arabic grammar. The detailed analysis of these benchmarks, which will be discussed in subsequent sections, provides clear evidence of how \textbf{GemmAr-7B-V1} is set to transform Arabic language processing in diverse applications.\newline
This paper is structured as follows: The Methodology section details the approaches used for dataset construction, including synthetic data generation and the use of the LoRA technique. The Analysis section presents benchmarks and discusses the results of tuning the model on Arabic NLP tasks. The final sections address critical aspects of our research and its broader impact. 
\section{Preliminaries}
\subsection{Instruction-tuning}
Instruction-tuning is a method designed to enhance the capabilities of pre-trained large language models (LLMs) by fine-tuning them with datasets composed of explicit natural language instructions and their corresponding responses \cite{81,82}. This technique aims to guide LLMs to better understand and respond to a variety of human requests, particularly those that include clear indications of the task to be performed \cite{83,84}. The practice of instruction-tuning can vary, including supervised learning with demonstrations or reinforcement learning from feedback data. However, supervised learning remains more common due to the scarcity of open resources for reinforcement learning-based approaches  \cite{selfinstruct,86}. \newline  In recent developments, publicly released foundation models have somewhat alleviated the high costs associated with training strong pre-trained language models. Nevertheless, these models often perform poorly in non-English languages, highlighting the need for more diverse linguistic datasets \cite{Llama, Llama2, mistral}. By using instruction datasets, models can generalize to new scenarios without dedicated retraining, allowing non-experts to interact with them naturally. The goal is to improve the model's capabilities, ensuring it maintains its high performance and delivers accurate responses in new linguistic contexts \cite{90}. \newline This makes instruction-tuning a critical step in improving the usability and accessibility of LLMs across different languages and tasks. In our project, we developed an Arabic language model by enhancing the performance of non-Arabic open-source models like Gemma-7B-IT. This was achieved by developing an Arabic instruction dataset \textbf{InstAr-500k} using both synthetic and real data.
\subsection{Instruction-datasets}
The instruction fine-tuning datasets are composed of paired textual data, wherein each pair consists of an "\textbf{\textit{instruction input}}" and a corresponding "\textbf{\textit{answer output}}". The "\textbf{\textit{instruction input}}" denotes the diverse range of requests or prompts issued by humans to the model, spanning multiple task types, including but not limited to classification, summarization, paraphrasing, and others. Conversely, the "\textbf{\textit{answer output}}" represents the model's generated responses that adhere to human expectations and align with the intended outcome of the original instruction.
\newline
There are generally two methods for creating instruction datasets: human-crafted datasets and LLM-generated datasets. Initially, humans created this type of dataset, but with the development of LLMs, it became possible to generate it using LLMs.

\underline{Human-Crafted Datasets:}\\ Human-crafted datasets are developed by individuals who follow specific rules and requirements to manually organize instructions. The creation process uses the deep intuitive understanding of language and context that human annotators possess, enabling precise interpretation of nuances and subjectivity. This iterative methodology produces high-quality, unique, and contextually rich datasets that enhance the performance of language models across various tasks. \newline These datasets appear in several forms: they may consist of annotated natural language data tailored for instruction output, like Flan \cite{52} and P3 \cite{6}, or may be completely new datasets created from scratch, as seen with Aya collection \cite{aya}.

\underline{Synthetic Datasets:} \\ Synthetic data, generated by algorithms rather than collected from real-world events, plays a pivotal role in training machine learning models where actual data may be scarce or sensitive. LLMs like GPT-3.5-Turbo \cite{gpt3} and GPT-4 \cite{gpt4} are particularly effective in creating high-quality synthetic datasets. These LLMs can simulate realistic and diverse data points by leveraging their deep learning capabilities. For example, datasets like InstructWild \cite{instructwild} and Self-Instruct \cite{selfinstruct} illustrate the application of LLMs in generating textual content that mimics human writing for NLP tasks.
\newline Additionally, the ability of LLMs to continuously learn and adapt ensures that the synthetic data remains relevant and reflective of evolving real-world conditions. This process is not only cost-effective but also speeds up the development cycle of machine learning models, making it a valuable tool across various domains and languages.

\section{Methodology}
\subsection{Overview}

To refine large language models for better performance in Arabic, we relied on a methodology that combines monolingual knowledge distillation \cite{100} and fine-tuning strategies across a spectrum of datasets—both synthetic and human-crafted to improve the performance of the Gemma-7B-IT model. 
\newline
The process began with the creation of a synthetic dataset using the Command R+ model\footnote{https://huggingface.co/CohereForAI/c4ai-command-r-plus}, designed to cover a wide range of tasks and contexts in Arabic. After collecting human-crafted datasets, we developed them further by performing some text pre-processing steps such as cleaning. This process ensured the datasets provided high-quality, contextually accurate instructions and responses. We then combined these datasets into a hybrid dataset, \textbf{InstAr-500k}, which balanced the strengths of both data types to form a rich training resource. \newline The fine-tuning process, conducted within the LLaMAFactory infrastructure \cite{llamafact}, primarily leverages the synthetic portion of the combined dataset for monolingual knowledge distillation \cite{100}, although it also incorporated the human-crafted dataset. This approach involved iterative adjustments to the models’ parameters to improve performance, ensuring a balanced improvement using both types of data.
\subsection{Training Data}
We offer in this section a detailed explanation of the dataset construction process, emphasizing the used methodologies and providing a quality analysis. 
The following Table \ref{tab:my_label} lists datasets used in our study, highlighting their origins and the range of tasks they cover. Each dataset was carefully selected to contribute to the effective training and fine-tuning of our Arabic language model.

\begin{table}[!ht]
    \centering
    \resizebox{0.8\columnwidth}{!}{%
    \begin{tabular}{|M{6.5cm}|M{2cm}|M{3cm}|M{2.2cm}|M{2cm}|}
    \specialrule{0.1em}{0em}{0em}
        \textbf{Datasets} &\textbf{Type}  &\textbf{ Tasks} & \textbf{N° of Samples} &\textbf{Samples \%} \\
        \hline
        Aya_Collection \cite{aya}&Human-crafted& Mixed &69068&14.35 \\
        \cline{1-1} \cline{3-5}
         ArabicaQA \cite{arabicaqa} &  & Open QA &61945  & 12.87\\
         \cline{1-1} \cline{3-5}
         CIDAR \cite{cidar}&  &  Mixed& 19986 & 4.15\\
         \cline{1-1} \cline{3-5}
         AQAD \cite{aqad}&  &  Open QA &17322  & 3.59\\
         \cline{1-1} \cline{3-5}
         Xtreme \cite{62}&  & Open QA  & 6926 & 1.43\\
         \cline{1-1} \cline{3-5}
        Ar_Math  \cite{Armath}  &  & Explanation & 6000 &1.24 \\
         \cline{1-1} \cline{3-5}
         Dawqas \cite{dawqas}&  &Open QA  &3209  &0.66 \\
         \cline{1-1} \cline{3-5}
         Ar_Medical \cite{57}& &Closed QA &1273 &0.26\\
         \cline{1-1} \cline{3-5}
         Arabic_RC \cite{55} & &Open QA &1003&0.20\\
         \hline Arabic_Categorization\tablefootnote{https://huggingface.co/datasets/abdalrahmanshahrour/arabic_categorization_data}_SANAD \cite{61} &Generated &Classification & 205540&42.70\\
         \cline{1-1} \cline{3-5}
         ClassicalAarabic_Poetry \cite{60}& &Closed QA &42650 &8.86\\
          \cline{1-1} \cline{3-5}
         101 Billion Arabic Words Dataset \cite{aloui2024101billionarabicwords}& & Mixed &8322 &1.71\\
         \cline{1-1} \cline{3-5}
         Abu_El_Khair \cite{abu}& &  Open QA&5990&1.24\\
         \cline{1-1} \cline{3-5}
        HTL_Ar_Sentiment \cite{arsentiments} & & Classification & 5000&1.03\\
        \cline{1-1} \cline{3-5}
         RES_Ar_Sentiment \cite{arsentiments}& &Classification &5000 &1.03\\
         \cline{1-1} \cline{3-5}
         BRAD \cite{bard}& &Extraction and Explanation&4682&0.97\\
         \cline{1-1} \cline{3-5}
         PROD_Ar_Sentiment  \cite{arsentiments}& &Classification & 4222&0.87\\
         \cline{1-1} \cline{3-5}
         xlel_wd_dictionary  \cite{xlel_wd_dictionary}& &Open QA & 2683&0.55\\
         \cline{1-1} \cline{3-5}
        Sahih_Al_Bukhari \cite{71} & &Explanation &2000 &0.41\\
        \cline{1-1} \cline{3-5}
         ABMC_Arabic_Corpus  \cite{abmc, 64, 65}& &Text Completion &1423 &0.29\\
         \cline{1-1} \cline{3-5}
        ARCD \cite{70}& &  Open QA&1382 &0.28\\
        \cline{1-1} \cline{3-5}
        MOV_Ar_Sentiment \cite{arsentiments}  & &Classification & 369&0.07\\
        \cline{1-1} \cline{3-5}
       Arabic_Text_ Summarization\tablefootnote{https://www.kaggle.com/datasets/abozekry/label-val}& & Summarization&154&0.03\\
       \hline
    \end{tabular}
    }
    \caption{Overview of different data sources for various tasks: A summary of key characteristics.}
    \label{tab:my_label}
\end{table}

\subsubsection{Instruction Dataset Construction}

\begin{figure}[!ht]
   \centering
  \includegraphics[width=0.8\textwidth]{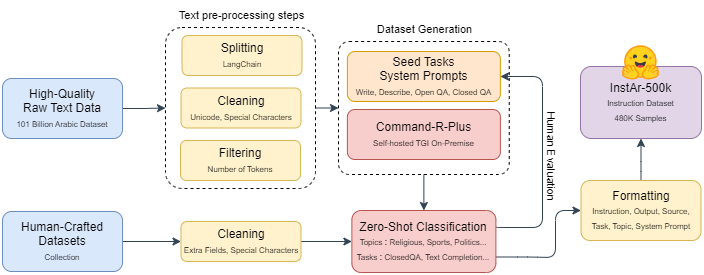}
   \caption{ Overview of the \textbf{InstAr-500k} dataset construction pipeline.}
   \label{filtered1}
 \end{figure}

Improvements in the chosen open-source LLMs for Arabic language understanding are centered on developing an extensive Arabic instruction dataset. The construction of our dataset is detailed through a pipeline, as illustrated in Figure \ref{filtered1}. \\
 \\
This section will explore the various stages of the dataset's construction, providing visual and descriptive insights into the different steps involved in preparing the data for the effective training of these models.

\begin{itemize}
\item \textbf{Human crafted Data: } \\
In our development of human-crafted data, we sourced existing datasets from prominent platforms such as Hugging Face, GitHub, and Kaggle, which were initially geared toward standard NLP tasks like classification. We then transformed these into instruction-response pairs to cultivate a varied instruction dataset. 
\newline This involved collecting raw text, classification labels, and other pertinent data, which we reformulated to fit instructional needs. \\Further, we conducted an extensive cleaning and filtering process to remove any irrelevant or low-quality data, ensuring the refined data was suitable for conversion into instruction-response pairs. During this phase, we standardized Arabic characters, numbers, and formatting, and adjusted the dataset based on token count to align with our fine-tuning specifications.

\item \textbf{Synthetic data: } \\
The construction of \textbf{InstAr-500k} dataset was a multi-stage process that used multiple tools and frameworks to ensure the production of high-quality data for fine-tuning our models, as illustrated in the previously mentioned pipeline. We began by using high-quality raw text data from the \textbf{101 Billion Arabic Words Dataset}  \cite{aloui2024101billionarabicwords}, specifically focusing on the Modern Standard Arabic (MSA) portion of the data, which served as the foundation for generating instruction-response pairs. \newline Using LangChain, we split this raw text into manageable segments to ensure coherence, followed by cleaning to remove unwanted characters, standardize Unicode, and address special characters. We also filtered the data based on token count to meet our fine-tuning specifications. To facilitate the generation of diverse and relevant data, we created seed task system prompts for tasks such as summarization, explanation, extraction, and open question answering (QA).  \newline Finally, we used the Command R+ model, hosted on self-managed HuggingFace TGI instances, to generate instruction-response pairs from the cleaned text. This model processed the seed prompts and produced a diverse set of instructions and corresponding responses, completing our dataset construction process. 
Refer to Appendices \ref{A}, \ref{B}, and \ref{C} for more details about the prompts used, contexts, and their outputs for three tasks: Open QA, Extraction, and Explanation.
\item \textbf{Data Combination: } \\
After cleaning both the human-crafted and generated datasets, we integrated them into a unified framework, where we further enhanced the dataset's consistency through zero-shot classification. This technique enabled us to classify the instructions by topics, such as politics or sports, and by task types, such as Open QA or Explanation. Subsequent human evaluation ensured that the data remained both consistent and diverse. Relying on feedback from this evaluation, we repeated the classification as necessary to further enrich the dataset’s diversity.  \newline Following this, we reformulated the data into a standardized format that includes key elements such as \textit{the instruction} (the task or query, expressed in Arabic), \textit{the expected output} (the response to the instruction), \textit{the source} (origin of the data),\textit{ the task} (specific nature of the task), \textit{the topic} (broader subject area of the instruction), and \textit{system prompts} (specific prompts guiding the system in generating responses). This formatting step ensured that all data adhered to a standardized structure (see Figure \ref{sample1}).
\newpage

\begin{figure}[!ht]
   \centering
  \includegraphics[width=1.01\textwidth]{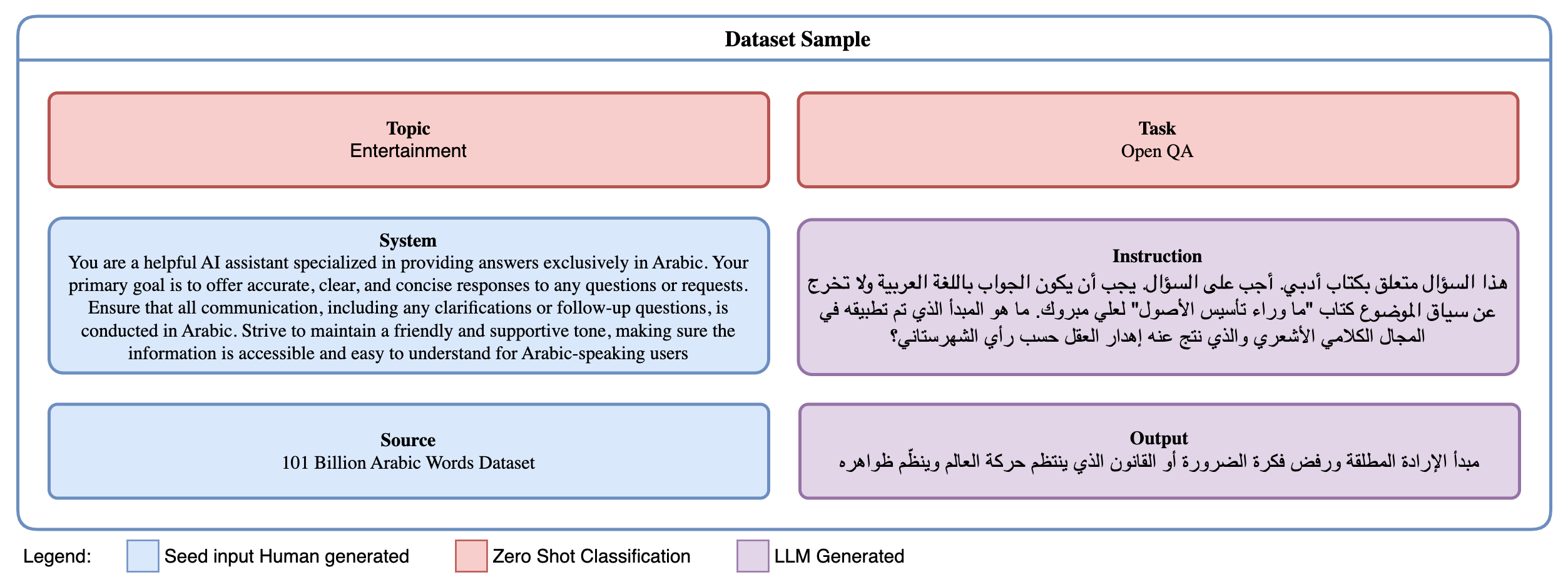}
   \caption{Sample from \textbf{InstAr-500k} dataset showcasing the standardized formatting.}
   \label{sample1}
 \end{figure}
 \end{itemize}

 \subsubsection{Quality analysis}
The \textbf{InstAr-500k} dataset includes a diverse range of tasks and sources, offering an examination of its scope and content. The high quality of the dataset results from multiple rounds of prompt engineering and detailed human evaluations to ensure clarity, relevance, and accuracy. Additionally, using high-quality raw text as the context in the synthetic data generation process has significantly improved the overall quality of the dataset.

 \begin{itemize}
\item \textbf{Token Length Distributions:}
\newline
\begin{figure}[!ht]
    \centering
    \includegraphics[width=0.6\linewidth]{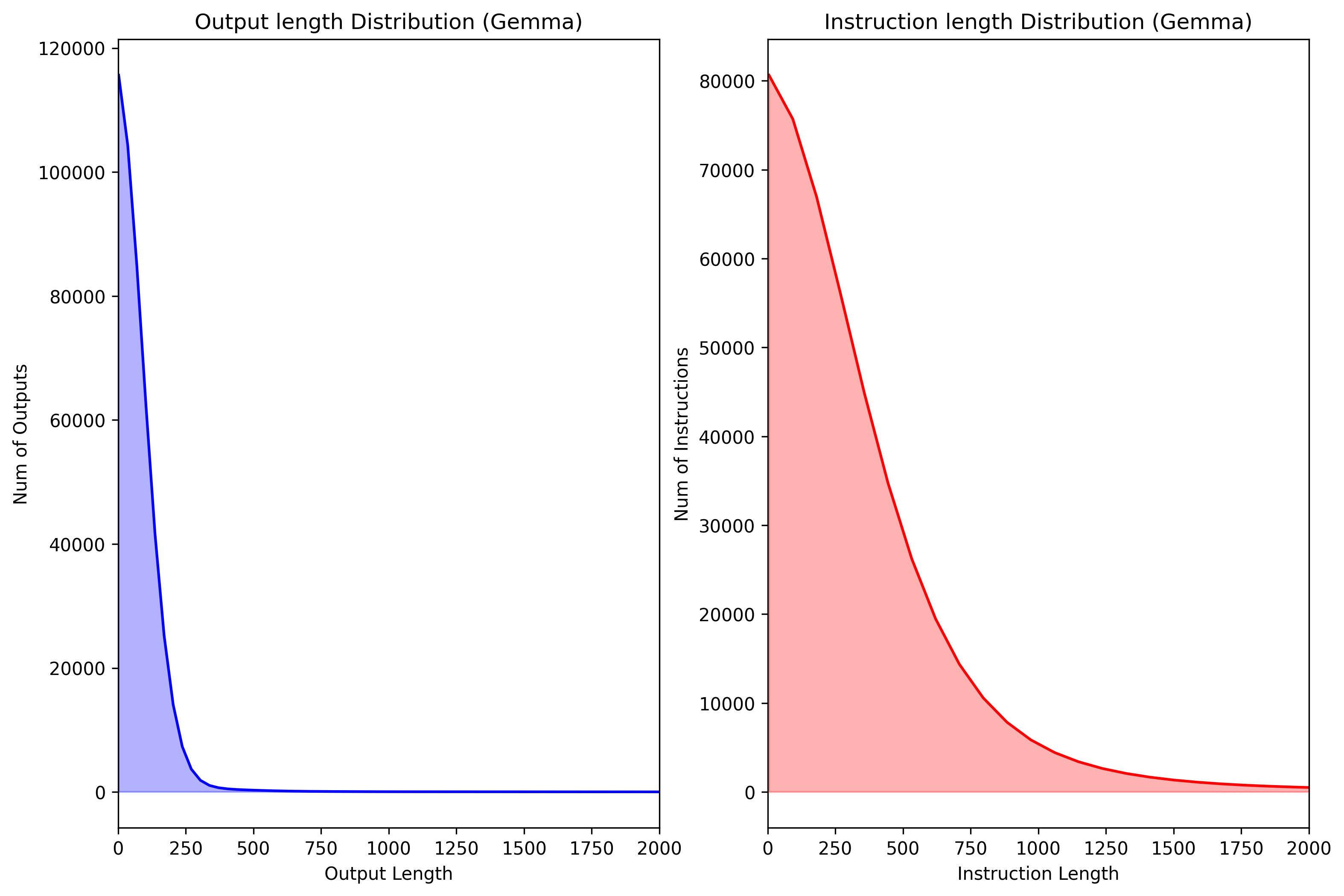}
    \caption{Tokens distribution in instructions and outputs for Gemma tokenizer.}
    \label{token}
\end{figure}

Figure \ref{token} illustrates the token length distributions for Gemma tokenizer, providing insights into the dataset's characteristics. The distributions highlight the dataset’s extensive scope, with 333,886,144 tokens processed in the inputs and 24,139,403 tokens generated, showcasing the dataset’s productive output. The logarithmic scaling reveals concentrated clusters of outputs and instructions within specific length ranges from 0 to 2,000 tokens, emphasizing the tokenizer’s impact on segmentation in computational linguistics.

 \item \textbf{Categorical Variety and Zero-Shot Classification:}
 \newline
The \textbf{InstAr-500k} dataset contains a wide range of categories with a well-balanced distribution across the following areas: Religion, Sports, Politics, Science \& Technology, Economy \& Finance, Entertainment, History, Health, Geography, and Travel. We used zero-shot classification to assign labels to instructions. We sourced these candidate labels for the zero-shot algorithm from the insights provided in the study referenced in \cite{76}. This approach enabled the model to generalize to new, unseen categories based on its acquired knowledge, thus eliminating the time-consuming and subjective process of manual labeling.
\item \textbf{Task Variety:}

The \textbf{InstAr-500k} dataset includes a diverse array of tasks: Classification, Open QA, Closed QA, Text Completion, Explanation, Brainstorming, Rewrite, Extraction \& Explanation, Generation, Extraction, and Summarization. We applied zero-shot classification to datasets that initially had mixed tasks, such as CIDAR \cite{cidar} and the Aya Collection \cite{aya}, to identify the specific task for each instruction accurately. This approach ensured the dataset's broad relevance across multiple domains.

\begin{figure}[h!]
    \centering
    \begin{minipage}{0.60\textwidth}
        \includegraphics[width= \linewidth]{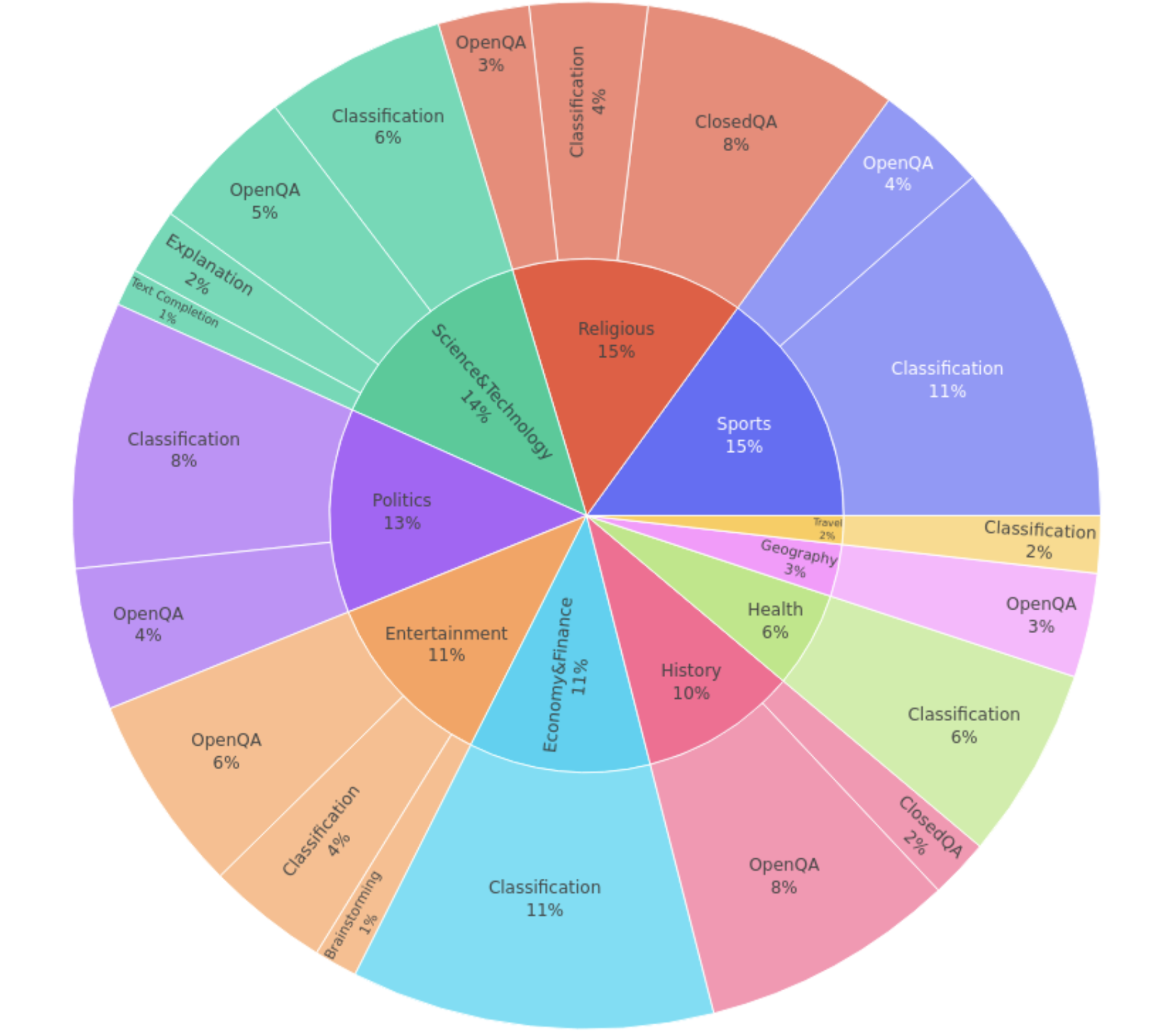}
        \caption{Distribution of tasks by topics.}
        \label{pie}
    \end{minipage}
    \begin{minipage}{0.30\textwidth}
        Figure \ref{pie} presents the distribution of tasks by topic, including only topics with percentages greater than 2\% for clarity.
    \end{minipage}
\end{figure}

\end{itemize}
 
 \subsection{Fine-tuning}
 In this subsection, we provide the technique and the parameters employed for supervised fine-tuning (SFT) within the LLaMAFactory framework \cite{llamafact}. For this, we used the Gemma 7B-IT model. Table  \ref{hyperpara} details its architectural parameters, including the number of layers, hidden dimensions, and other metrics.


\begin{table} [!ht]
\centering
\resizebox{0.95\textwidth}{!}{%
\begin{tabular}{|M{2cm}|M{1.7cm}|M{1.3cm}|M{2.3cm}|M{1.6cm}|M{2cm}|M{1.6cm}|M{1.4cm}|M{2cm}|M{2.2cm}|}
\specialrule{0.1em}{0em}{0em}
& \textbf{D-Model} &\textbf{Layers}&\textbf{FFN Hidden Dims.}&\textbf{N° heads}&\textbf{N° KV heads}&\textbf{Head Size}&\textbf{Vocab size}&\textbf{Embedding Params.}&\textbf{Non-embedding Params.}\\
\hline
\textbf{Gemma 7B-IT}&3072&28&49152&16&16&256&256128&786.825.216&7.751.248.896\\
\hline
\end{tabular}
}
\caption{Model architecture parameters. }
\label{hyperpara}
\end{table}

In addition to model architecture, the choice of hyperparameters has a prominent role in the fine-tuning process, directly influencing the model’s output. During the fine-tuning of the Gemma 7B-IT model, specific hyperparameters were selected to optimize training. Table \ref{FTP}  summarizes the key hyperparameters, highlighting the calibration necessary for the model training.

\begin{table} [!ht]
\centering
\resizebox{0.9\textwidth}{!}{%
\begin{tabular}{|M{3.5cm}|M{2cm}|M{2cm}|M{2cm}|M{2.3cm}|M{2cm}|M{2cm}|}
\specialrule{0.1em}{0em}{0em}

\textbf{Model}&\textbf{Learning Rate}&\textbf{Batch Size}&\textbf{Epochs}&\textbf{Gradient Accumulation Steps}&\textbf{Gradient Norm}&\textbf{Cutoff Length}\\
\hline
\textbf{Gemma 7B-IT}&1e-4&2.0&3&16&2.0&2048\\
\hline
\end{tabular}
}
\caption{Fine-tuning hyperparameters.}
\label{FTP}
\end{table}

To further enhance the model’s performance, we integrated a range of advanced configurations and techniques within our fine-tuning process: 
\begin{itemize}
   
  \item \textbf{RoPE:} We used dynamic Rotary Positional Embeddings (RoPE) \cite{rope} to improve long context extrapolation and enhance performance on downstream tasks with short context lengths. This dynamic RoPE scaling effectively manages longer sequences, offering superior performance across different context lengths.

  \item \textbf{Flash Attention: }To further optimize our fine-tuning process, we used Flash Attention \cite{flashAtt} (flash\_attn2) as a booster. This improves memory efficiency and computational speed, enabling us to manage larger batches and longer sequences more effectively

  \item \textbf{Learning Rate Scheduler:} We implemented a cosine learning rate scheduler. This approach helped us gradually reduce the learning rate over time, ensuring smoother convergence and preventing abrupt changes that could destabilize the training process.

  \item \textbf{Optimizer:} We employed the AdamW\_torch optimizer \cite{adamw}, which combines the Adam optimization algorithm with weight decay correction. This choice helped us to maintain efficient and stable training by preventing overfitting and ensuring better generalization.

  \item \textbf{Precision:} We used bfloat16 (bf16) precision during training. This allowed us to faster compute and reduce memory usage without significantly sacrificing model accuracy, enhancing overall training efficiency

  \item \textbf{LoRA:} In our fine-tuning process, we used LoRA (Low-Rank Adaptation) \cite{lora} to efficiently adapt models for Arabic language tasks. LoRA helped us reduce the number of trainable parameters, making the fine-tuning process more efficient without compromising performance. 
 \end{itemize}
 \begin{table} [!ht]
\centering
\begin{tabular}{|c|c|c|c|}
\specialrule{0.1em}{0em}{0em}
\textbf{Rank} &\textbf{Alpha}&\textbf{Dropout}&\textbf{Target Layers}\\
\hline
8&16&0.01&All\\
\hline

\end{tabular}
\caption{LoRA configuration parameters.}
\label{loraconfig}
\end{table}

\textbf{Impact of hyperparameters fine-tuning}

In our efforts to boost the performance of Gemma-7B-IT, we refined quantization techniques by reducing the precision of model parameters. This adjustment enabled the model to prioritize important features during training, resulting in more efficient memory usage and faster training times.\newline Implementing a warm-up step was crucial for optimization. We improved model convergence by starting with a lower learning rate and gradually increasing it. This gradual warm-up established the model's foundation, leading to smoother training and enhanced overall performance. Additionally, changing the learning rate to smaller values ensured gradual and precise updates to the model's parameters, preventing irregular behavior.\newline
We also exposed the model to a larger and more diverse dataset by increasing the number of training examples. This strategy boosted the model's generalization ability and reduced overfitting, making it adaptable to various scenarios. For specific configurations of LoRA, we defined precise relationships between hyperparameters. For example, setting the alpha $(\alpha) \text{ based on the rank } (r) (\alpha=2*r; r=8 \text{ and } \alpha=16)$ helped in the exploration-exploitation trade-off, allowing the model to escape local minima and discover optimal solutions.\newline
We adjusted the cutoff value during parameter trimming to allow the model to capture complex patterns effectively. By retaining a larger set of parameters, the model handled fine-tuning tasks better and understood complex information, balancing general and specific pattern recognition.\newline By increasing the frequency of updates, or save steps, from 100 to 1000, we achieved a more controlled learning process. This change allowed the model's parameters to be adjusted and evaluated more frequently, enhancing convergence and stability. The synchronized approach of maintaining equal save and eval steps ensured effective use of computational resources and prompt issue identification. \newline
Finally, Max Gradient Normalization reinforced training stability by normalizing gradients to a maximum norm. This adjustment prevented extreme gradient values, especially when dealing with outliers or noisy data. Our journey of optimizing LLMs has led to significant performance improvements.

 \subsection{Environment}
 \subsubsection{Dataset generation environment}
 The dataset production environment included several key components.  We used Text Generation Inference (TGI) from Hugging Face for Command R+ inference, deployed to an on-premises Kubernetes cluster on a custom-built Nvidia HGX 8xL40S cluster, allowing for efficient management through a Helm chart.
The deployment leveraged GPU resources to accelerate the inference process and ensure more efficient handling of large datasets. We used Jupyter Lab with Python version 3.12 for development and text editing.

 \subsubsection{Fine Tuning environment}
 We used the Azure AI platform to create a solid environment for our Natural Language Generation (NLG) model development and testing. We chose the Standard\_NC96ads\_A100\_v4 instance type, equipped with 4 x NVIDIA A100 GPUs. This selection provided us with the necessary computational resources to handle large-scale data processing and model training.
 \begin{itemize}
 \item \textbf{Configuration and Software: } \\
 After setting up the infrastructure, we configured the necessary NVIDIA drivers and CUDA to optimize the GPU performance. We then installed JupyterLab to enable testing for efficient testing and iterative development of our synthetic data generation process. We used the HuggingFace text generation inference Docker image for rapid inference and text generation. This pre-built image provided a ready-to-use environment with all necessary dependencies and libraries, allowing us to focus on model development without spending time on manual setup.
\item \textbf{Model Evaluation:}\\
We tracked progress and ranked our model’s performance using the Open Arabic  LLM Evaluation Leaderboard (OALL) \cite{oall}. LLMs on the OALL are evaluated with LightEval, a unified framework from the Hugging Face, to test and assess causal language models across multiple evaluation tasks. This includes Arabic translations of benchmarks like MMLU \cite{39}, Exam \cite{40}, ARC-Challenge \cite{43}, ARC-Easy \cite{43}, BOOLQ \cite{44}, COPA \cite{45}, HellaSwag \cite{46}, OPENBOOK-QA \cite{47}, PIQA \cite{48}, RACE \cite{49}, SCIQ \cite{50}, and TOXIGEN \cite{51}. The leaderboard also features benchmarks specifically created for Arabic and its cultural context, such as AlGhafa \cite{alghafa} and ACVA \cite{11}.
\end{itemize}
\section{Analysis}
\subsection{Benchmarks}
Our study applied a set of benchmarks to evaluate the performance of our model across multiple domains. These benchmarks were selected to cover a broad range of tasks, ensuring a diverse assessment. Key components of our evaluation included the Arabic MMLU Benchmark, which is the translated version of MMLU (Massive Multi-task Language Understanding) provided by the OALL team\footnote{https://huggingface.co/OALL}. This benchmark is a standardized method for assessing AI performance on a range of tasks, from simple mathematics to complex legal reasoning in Arabic. It consists of 57 tasks across numerous domains, including elementary mathematics, history, computer science, and law, requiring models to demonstrate a broad knowledge base and problem-solving skills. \newline Additionally, we used the ACVA Benchmark (Arabic Cultural and Value Alignment), introduced by AceGPT \cite{11}, as a benchmark for evaluating our model's alignment with Arabic cultural nuances and values. This examination is crucial for understanding the model's adaptation to the unique linguistic and cultural context of the Arabic language. \newline
To broaden the evaluation spectrum, we included additional benchmarks such as the AlGhafa Benchmark \cite{alghafa}, developed by the TII LLM team\footnote{https://www.tii.ae/}, as well as Arabic-specific versions of benchmarks like Arabic-EXAMS and Arabic-ARC-Challenge. These benchmarks evaluate models on reading comprehension, sentiment analysis, and question answering, ensuring a reliable evaluation of the model's performance.

\subsection{Results}

The results from various models on the Open Arabic LLM Leaderboard (OALL)\footnote{https://huggingface.co/spaces/OALL/Open-Arabic-LLM-Leaderboard} reveal diverse performances across multiple Arabic Natural Language Understanding  (NLU) tasks. Our fine-tuned model, GemmAr-7B-V1, exhibits strong performances with average scores of 47.27\%.
\newline
\begin{figure}[!ht]
    \centering
    \includegraphics[width= 0.8 \linewidth]{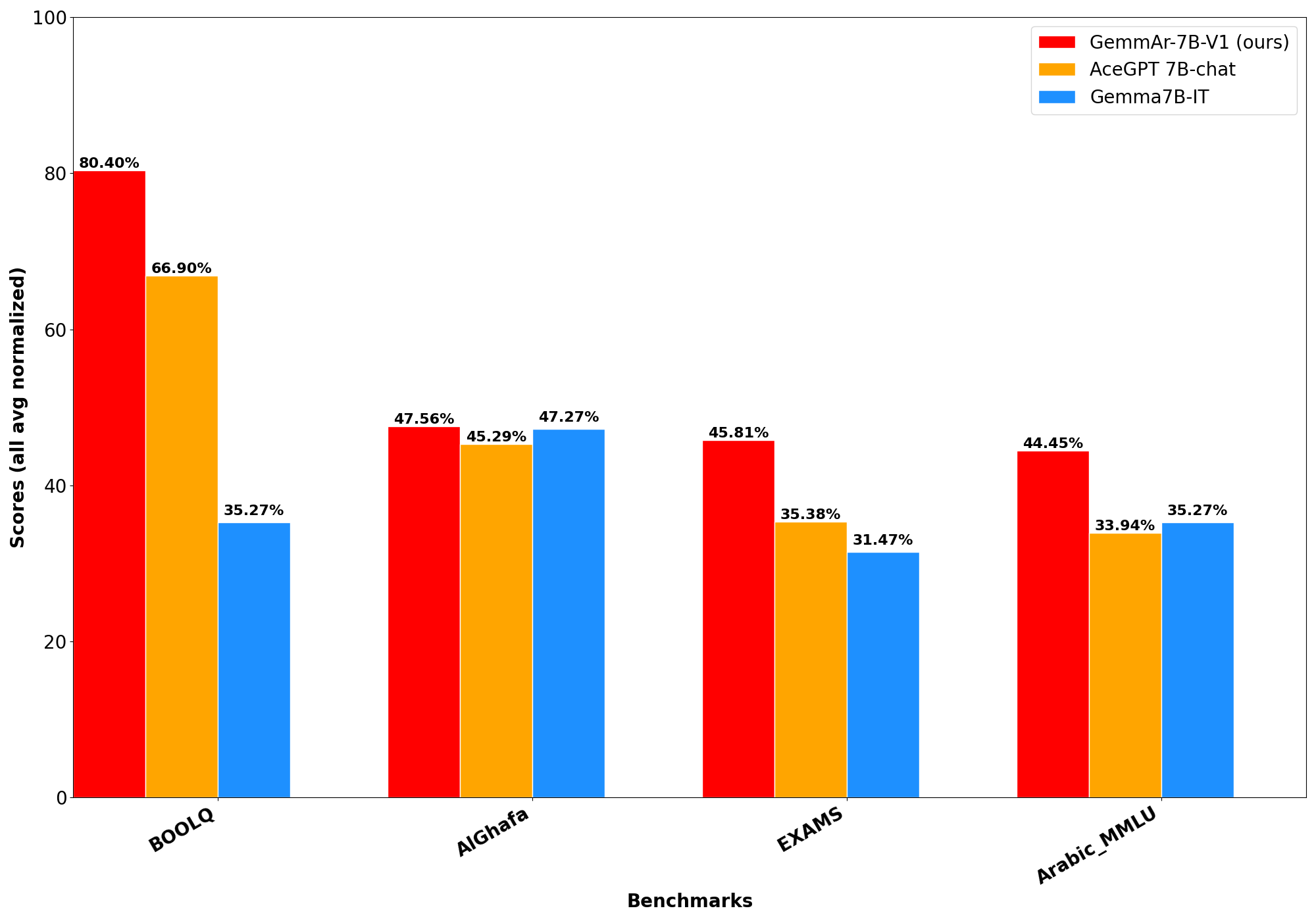}
    \caption{  Performance Scores Comparison of \textbf{GemmAr-7B-V1}, AceGPT-7B-chat, Gemma 7B-IT across different Benchmarks.}
    \label{fig:my_label}
\end{figure}
\\
Figure \ref{fig:my_label} illustrates the performance of several key benchmarks, including Arabic\_MMLU, AlGhafa, EXAMS, and BOOLQ. These benchmarks are compared against the base model, Gemma 7B-IT \cite{gemma}, and AceGPT 7B-chat \cite{11} on the OALL leaderboard. The results in Table \ref{tab:benchmark_scores}, highlight that our model, \textbf{GemmAr-7B-V1}, stands out as the best-performing model, with higher scores for Arabic tasks. It demonstrates strong performance in the Arabic\_MMLU, AlGhafa, and Toxigen\_Ar benchmarks, which evaluate multi-tasking abilities, contextual understanding, and vocabulary.
It also outperforms the other models in most benchmarks, particularly in the SCIQ\_Ar and RACE\_Ar, which assess commonsense reasoning and reading comprehension respectively.

\begin{table}[!ht]
    \centering
    \resizebox{0.8\textwidth}{!}{%
    \begin{tabular}{|M{4cm}|M{3.5cm}|M{3cm}|M{3cm}|}
    \hline    
    &\multicolumn{3}{|c|}{\textbf{Benchmark Scores per model}}\\ \hline
    \textbf{Benchmarks} & \textbf{GemmAr-7B-V1 (ours)} & \textbf{Gemma7B-IT}& \textbf{AceGPT 7B-chat}\\ \hline
    Arabic_MMLU & \textbf{44.45}  & 35.27 &33.94 \\ \hline
    ACVA & 42.76 & 40.29 & \textbf{46.84}\\ \hline
    AlGhafa & \textbf{47.56} & 47.27& 45.29\\ \hline
    Arabic_Exams & \textbf{45.81}& 31.47 & 35.38\\ \hline
    ARC_Challenge & \textbf{38.36}& 34.05&32.16\\ \hline
    ARC_Easy &\textbf{38.54 } & 35.07 &34.39\\ \hline
    Boolq_Ar & \textbf{80.40} &59.69&66.90\\ \hline
    Copa_ext_Ar & 48.88&51.11&\textbf{55.56}\\ \hline
    Hellaswag_Ar & \textbf{26.65} &26.42 &26.54\\ \hline
    OpenBook_QA &40.61 & \textbf{41.41} &38.59 \\ \hline
    Piqa_Ar &\textbf{ 57.06} &54.88& 54.45\\ \hline
    Race_Ar & \textbf{35.46} & 35.20 &33.15\\ \hline
    Sciq_Ar & \textbf{50.35 } & 40.80& 44.72\\ \hline
    Toxigen_Ar & \textbf{64.81}&57.22 &43.10\\ \specialrule{0.1em}{0em}{0em}
    Total Average score normalized & \textbf{47.27}& 42.15 &42.21\\ \hline
    \end{tabular}}
    \caption{OALL evalset 0-shot benchmark scores.}
    \label{tab:benchmark_scores}
\end{table}

\section{Related Work}
Significant efforts have focused on diversifying instruction datasets, primarily focusing on the English language. These datasets can be categorized into two main types: those generated by Large Language Models (LLMs) and those created by humans using templates. Examples of LLM-generated datasets include Stanford Alpaca \cite{1}, Databricks' Dolly \cite{2}, and SELF-INSTRUCT \cite{selfinstruct}. Auto-Instruct \cite{25} aims to improve instruction quality for LLMs by leveraging their generative abilities to produce multiple instructions, which are then ranked by a scoring model trained on 575 NLP tasks. In contrast, human-crafted instruction datasets use templates to ensure consistency and coverage of various instruction types. Prominent examples include P3 \cite{6} and NATURAL INSTRUCTIONS \cite{82}, which focus on natural language processing tasks. Although the majority of research has focused on the English language, there have been significant contributions in Arabic. CIDAR \cite{cidar} was the first open Arabic instruction-tuning dataset, and the Aya Collection \cite{aya} offers a dataset in 101 languages, including Arabic.\\
Fine-tuning large language models presents several challenges, such as potential knowledge erosion, where modifying all parameters can lead to forgetting previously learned tasks \cite{27}. To address this issue, \cite{80} introduced Parameter Efficient Fine-Tuning (PEFT), which edits only a subset of parameters to help retain previously acquired knowledge more effectively \cite{28}. One such PEFT technique, Low-Rank Adaptation of Large Language Models (LoRA), proposed by \cite{lora}, reduces the number of trainable parameters, optimizing the fine-tuning process. By applying LoRA to pre-trained models, high performance can be achieved with minimal computational costs.\\
Regarding LLM evaluations, numerous frameworks have been proposed. HELM \cite{42} and LM Evaluation Harness \cite{41} cover a broad range of NLP tasks but typically focus on assessing base models rather than instruction-tuned ones. Newer evaluation frameworks, such as AlpacaEval and Chat Bot Arena, evaluate the open-ended instruction-following abilities of LMs. These frameworks use other models (AlpacaEval \cite{36}) or humans (Chatbot Arena \cite{37}) as annotators to assess the models' outputs. Another way of assessing LLMs is the Open Multilingual LLM Evaluation Leaderboard \cite{35}, which tracks progress and ranks the performance of LLMs across different languages, including Arabic. LLMs on the OALL are evaluated with LightEval, a unified framework from the Hugging Face Eval Team, to test and assess causal language models across multiple evaluation tasks. They translated benchmarks to Arabic, such as MMLU \cite{39}, Exam \cite{40}, ARC-Challenge \cite{43}, ARC-Easy \cite{43}, BOOLQ \cite{44}, COPA \cite{45}, HellaSwag \cite{46}, OPENBOOK-QA \cite{47}, PIQA \cite{48}, RACE \cite{49}, SCIQ \cite{50}, and TOXIGEN \cite{51}. The leaderboard also includes benchmarks specifically created for the Arabic language and its cultural alignments, such as AlGhafa \cite{alghafa} and ACVA \cite{11}.

\section{Conclusion}
In our continuous pursuit of improving Arabic NLP, we are delighted to introduce \textbf{GemmAr-7B-V1}, a  fine-tuned version of a Large Language Model specifically developed for Arabic using our crafted Arabic-instructed dataset \textbf{InstAr-500k}. By combining the strength of the LLM with our customized training approach, we aim to remove any barrier that hinders the performance of Arabic language tasks and make \textbf{GemmAr-7B-V1} trusted companions for researchers and Arabic LLM enthusiasts. 
We are excited about the potential impact of this model and look forward to witnessing the different use cases that will be built on \textbf{GemmAr-7B-V1}.

\section*{Limitations}
Despite the promising outcomes we have achieved, several limitations need to be addressed.\newline  Firstly, hardware constraints limited our ability to experiment with alternative parameter settings, particularly those affecting GPU memory.\newline 
Secondly, although the dataset's diversity represents an improvement over previous versions, it could be further explored in other tasks like Brainstorming and Role Playing. \newline Thirdly, the dataset currently includes only Modern Standard Arabic (MSA) instructions and lacks dialectal variations. This restricts its applicability to various regions. \newline 
Moreover, the evaluation metrics display a Western-centric bias, with subtopics like US History and European History potentially affecting the relevance of our findings across different contexts. As researchers, we acknowledge the need for continuous expansion and refinement of resources, addressing both technical feasibility and cross-cultural representation. This recognition forms the foundation for future efforts aimed at advancing Arabic NLP and ensuring equitable access to cutting-edge technologies. 

\section*{Ethical Considerations}
In this paper, we applied a new finetuning approach for two Large Language Models (LLMs) using \textbf{InstAr-500k} dataset. Our goal was to enhance the performance and adaptability of these models for Arabic speakers while being mindful of ethical implications and striving for responsible practices.\newline
We addressed bias and fairness by meticulously curating and auditing the data to ensure cultural sensitivity, diversity, and inclusivity. This helped promote equitable representations and reduce potential biases in the models' responses. Additionally, We prioritized user privacy and data protection, as our finetuning methodology did not involve collecting or storing any personally identifiable information. We constructed the Arabic instruction dataset using synthetic data, anonymized content, or data obtained with informed consent, adhering to secure data handling practices and relevant data protection regulations.\newline
We upheld transparency and accountability by disclosing the capabilities and limitations of our fine-tuned models, as well as any potential risks associated with their use. This included clear explanations of the models' evaluation, the Arabic-instructed dataset used, and any known limitations or biases specific to the Arabic language or cultural context. By addressing these ethical considerations, we aim to contribute to the responsible development and deployment of LLMs for Arabic chat applications, ensuring the protection of user privacy, the promotion of accurate and reliable information, and the alignment of models with cultural values and norms in Arabic-speaking societies.

\bibliography{main}
\newpage

\appendix
\section{\textbf{Open QA}}\label{A}
\begin{figure}[!ht]
   \centering
   \includegraphics[width=0.75\textwidth]{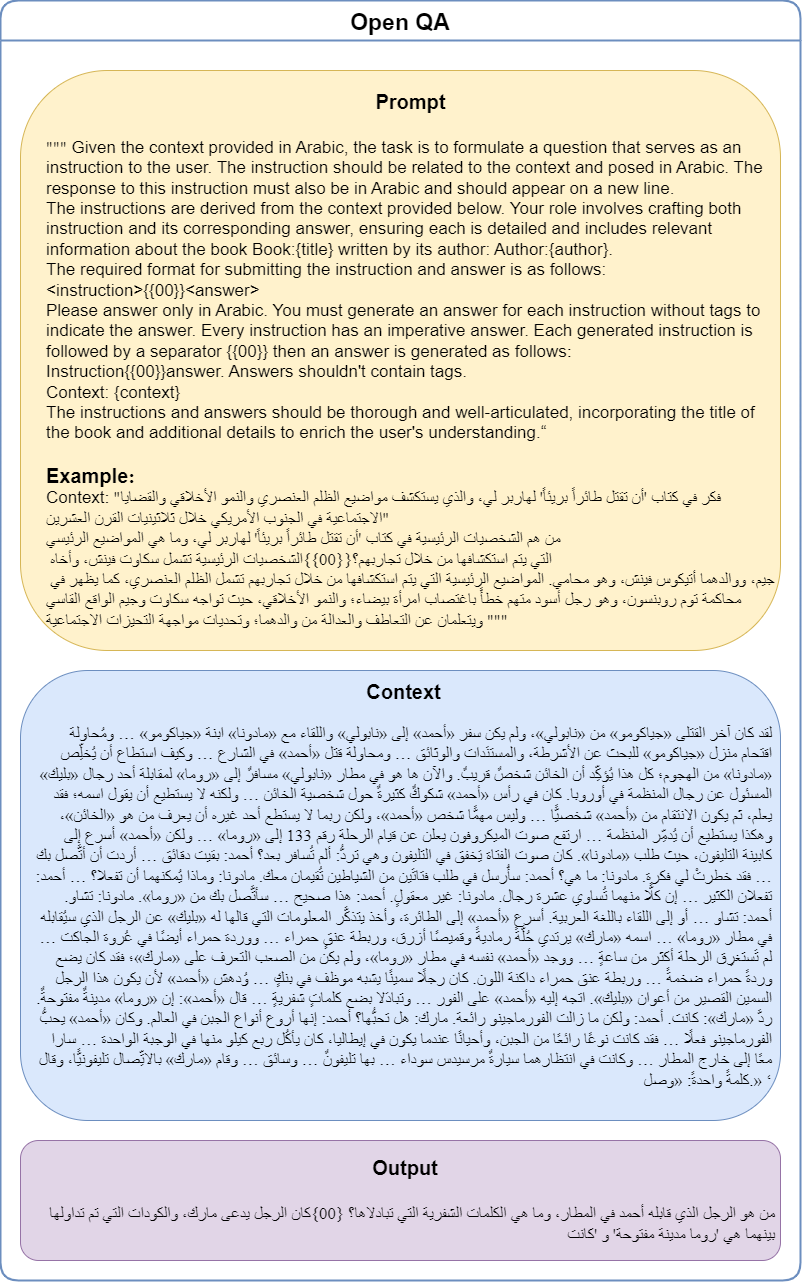}
   \caption{Example of the prompt, context, and output for the Open QA task.}
   \label{fig:openQA}
 \end{figure}
\newpage
\section{\textbf{Extraction}}\label{B}
\begin{figure}[!ht]
   \centering
   \includegraphics[width=0.85\textwidth]{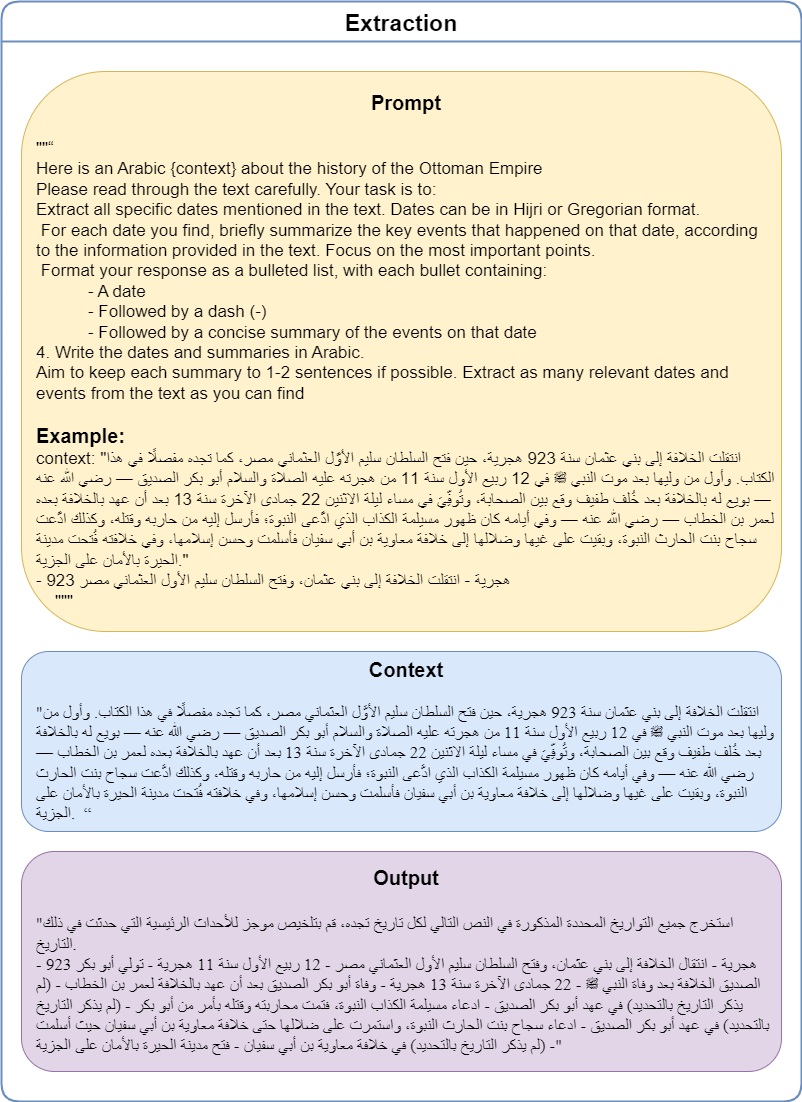}
   \caption{Example of the prompt, context, and output for the Extraction task.}
   \label{fig:extraction}
 \end{figure}
\newpage
\section{\textbf{Explanation}}\label{C}
\begin{figure}[!ht]
   \centering
   \includegraphics[width=0.8\textwidth]{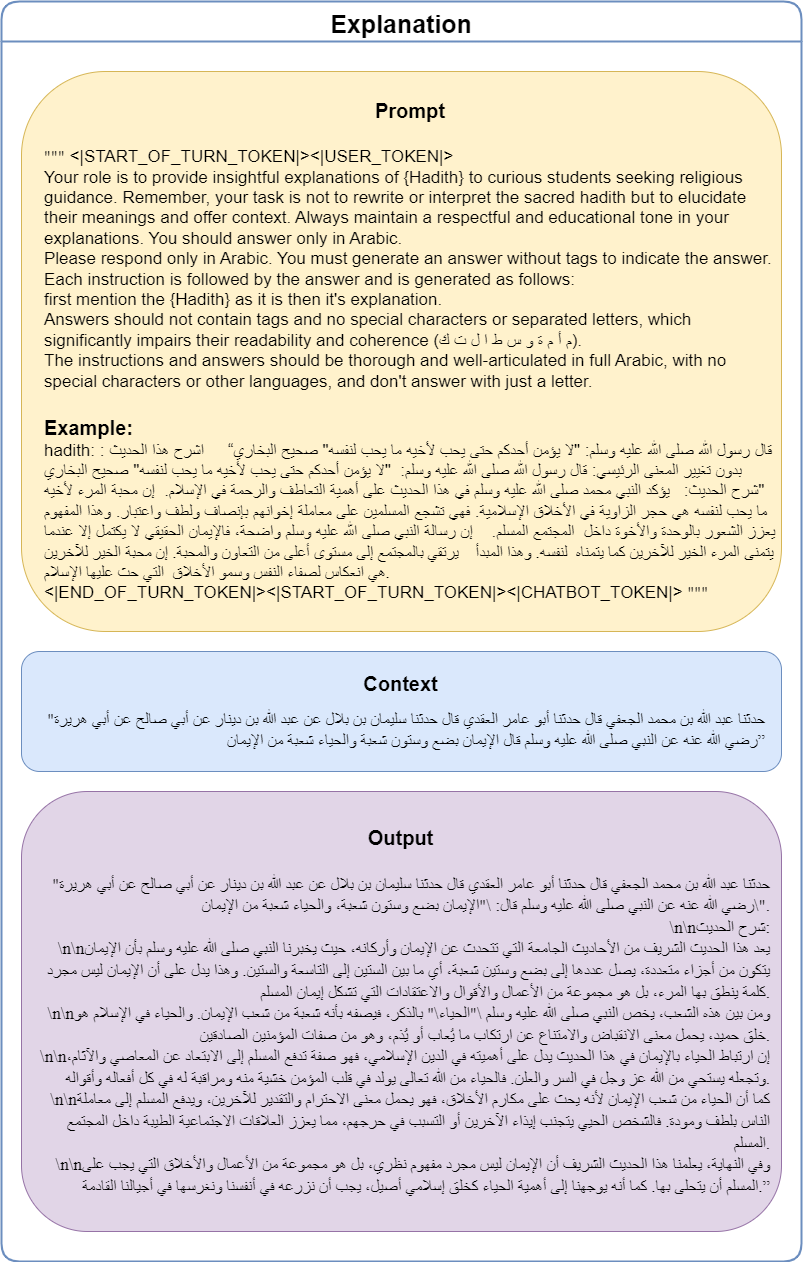}
   \caption{Example of the prompt, context, and output for the Explanation task.}
   \label{fig:explanation}
 \end{figure}

\end{document}